%% file: main.tex
\documentclass{article}

\PassOptionsToPackage{numbers, compress}{natbib}

\usepackage[preprint]{neurips_2026}

\usepackage[utf8]{inputenc} 
\usepackage[T1]{fontenc}    
\usepackage{hyperref}       
\usepackage{url}            
\usepackage{booktabs}       
\usepackage{amsfonts}       
\usepackage{nicefrac}       
\usepackage{microtype}      
\usepackage{xcolor}         

\usepackage{graphicx}
\usepackage{amssymb}
\usepackage{colortbl}
\usepackage{enumitem}
\usepackage{multirow}
\usepackage{algorithm}
\usepackage{algorithmic}
\usepackage{siunitx}
\usepackage{caption}
\usepackage{subcaption}     

\title{VideoRouter: Query-Adaptive Dual Routing for Efficient Long-Video Understanding}

\author{%
Kuanwei Lin\thanks{Equal contribution.},
 \quad Wenhao Zhang\footnotemark[1],
 \quad Ge Li\thanks{Corresponding author.} \\
School of Electronic and Computer Engineering, Peking University \\
\texttt{\{gwlin, whzhang25\}@stu.pku.edu.cn} \\
\texttt{geli@ece.pku.edu.cn}
}

\begin{document}

\maketitle

\input{sections/abstract}
\input{sections/01_introduction}
\input{sections/02_related_work}
\input{sections/03_method}
\input{sections/04_supervision_data}
\input{sections/05_experiments}

\input{sections/06_conclusion}

\clearpage
\bibliographystyle{unsrt} 
\bibliography{main}

\clearpage
\appendix
\input{appendix}


\end{document}

%% file: sections/abstract.tex
\begin{abstract}
Video large multimodal models increasingly face a scalability bottleneck: long videos produce excessively long visual-token sequences, which sharply increase memory and latency during inference. While existing compression methods are effective in specific settings, most are either weakly query-aware or apply a fixed compression policy across frames, proving suboptimal when visual evidence is unevenly distributed over time. To address this, we present \textbf{VideoRouter}, a query-adaptive dual-router framework built on InternVL for budgeted evidence allocation. The Semantic Router predicts the dominant allocation policy, choosing between broad temporal coverage and adaptive high-resolution preservation, while the Image Router uses early LLM layers to score frame relevance. This enables aggressive compression on less relevant frames while preserving detail on critical evidence frames. To train both routers, we build Video-QTR-10K for allocation-policy supervision and Video-FLR-200K for frame-relevance supervision. Experiments on VideoMME, MLVU, and LongVideoBench show that VideoRouter consistently improves over the InternVL baseline under comparable or lower budgets, achieving up to a 67.9\% token reduction.
\end{abstract}

%% file: sections/01_introduction.tex
\section{Introduction}
\label{sec:intro}

Video understanding with Multimodal Large Language Models (MLLMs)~\cite{videochatgpt,videollava,chatunivi,llavanextvideo,pllava} is fundamentally bottlenecked by the massive number of visual tokens generated for long videos, causing severe memory and latency issues~\cite{llavavideo}. Existing token-compression methods~\cite{prumerge,visionzip,fastv,sparsevlm,metok,dtoma} reduce this cost, but most of them allocate the same compression policy across frames or make token decisions before sufficient question-conditioned reasoning. This is suboptimal because long videos contain highly non-uniform evidence: some questions require broad temporal coverage, while others depend on a few spatially detailed moments, and many practical queries mix both needs.

\begin{figure}[t!]
\centering
\includegraphics[width=\textwidth]{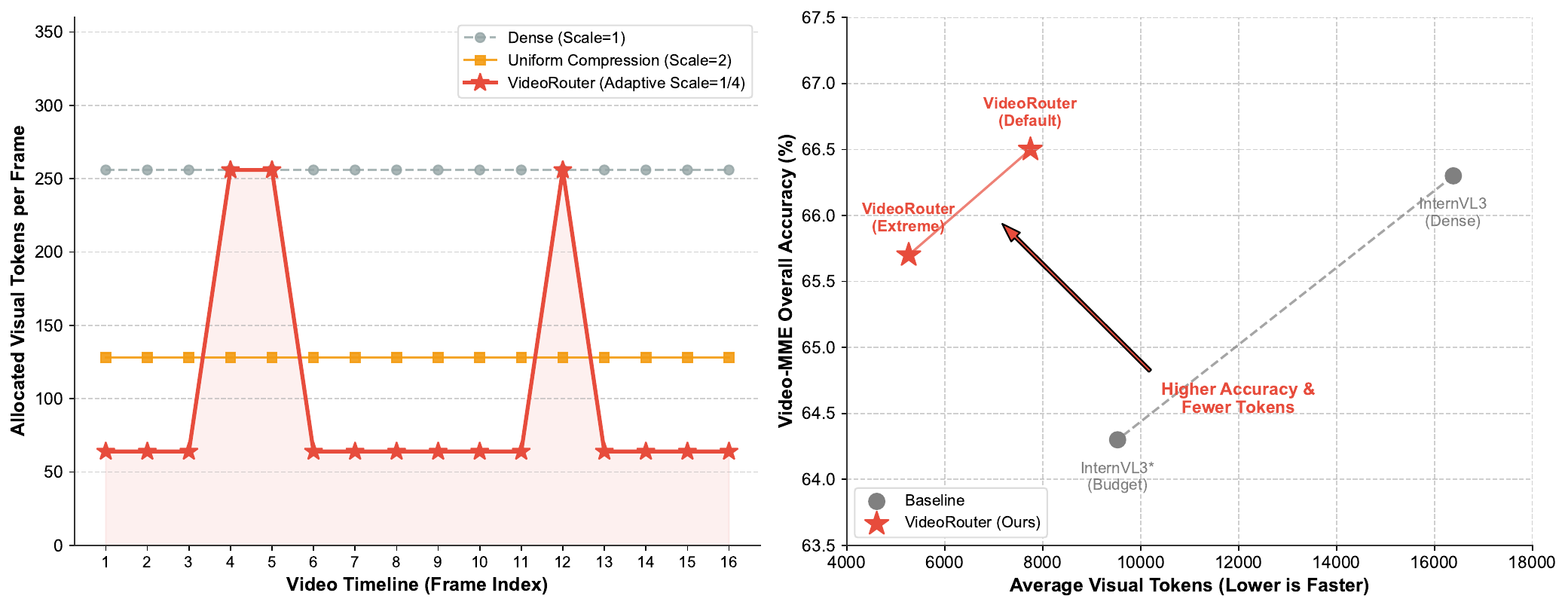}
\caption{\textbf{Left:} Frame-level dynamic token allocation by VideoRouter compared to dense and uniform baselines. \textbf{Right:} Accuracy vs. efficiency trade-off on Video-MME, demonstrating higher performance with significantly fewer visual tokens.}
\label{fig:intro_teaser}
\end{figure}

To address these issues, we frame long-video compression as \emph{budgeted evidence allocation}: given a limited visual-token budget, the model should decide whether to spend tokens broadly for temporal coverage or concentrate them on high-evidence frames. Importantly, our routing labels should not be interpreted as a rigid taxonomy of user intent. Instead, they define the dominant allocation policy under a fixed budget: a \emph{Global} policy preserves coarse coverage over the full video, while a \emph{Fragment} policy enables non-uniform high-resolution allocation over one or multiple evidence-bearing events.

Motivated by this view, we propose \textbf{VideoRouter}, a query-adaptive dual-router framework built upon InternVL~\cite{internvl}. A Semantic Router first predicts the dominant allocation policy, selecting between uniform coverage and adaptive allocation. When adaptive allocation is selected, an Image Router reuses early multimodal LLM layers to estimate frame-query relevance and drive fine-grained token assignment. The two routers are complementary: the Semantic Router prevents over-focusing on holistic queries, while the Image Router prevents uniform compression from destroying localized evidence. Together, they dynamically retain high resolution for critical frames, preserve coarse context for less relevant frames, and operate under an explicit token budget.

To supervise this routing behavior, we construct two specialized datasets. Video-QTR-10K provides 10,000 video-query pairs labeled by the preferred allocation policy, while Video-FLR-200K contains 200,664 QA pairs with frame-level relevance annotations. These datasets are not additional task-specific test knowledge; they provide supervision for deciding how to allocate visual tokens before answer generation.

Our key contributions are summarized as follows:
\begin{enumerate}[leftmargin=2em]
    \item \textbf{Budgeted Evidence Allocation}: We formulate long-video token compression as an explicit budgeted allocation problem, where the model chooses between broad temporal coverage and adaptive high-resolution preservation.
    \item \textbf{Dual-Router Architecture}: We introduce a query-adaptive framework that combines policy-level semantic routing with frame-level relevance estimation from early multimodal LLM layers.
    \item \textbf{Routing Supervision}: We present Video-QTR-10K and Video-FLR-200K, two dedicated datasets designed to supervise allocation-policy prediction and frame-query relevance estimation.
\end{enumerate}

%% file: sections/02_related_work.tex
\section{Related Work}
\label{sec:related}

Video token compression methods reduce the visual-token cost either before language modeling, during LLM inference, or through multi-stage pipelines. Pre-LLM approaches such as PruMerge~\cite{prumerge} and VisionZip~\cite{visionzip} inherit ideas from token dropping, merging, and adaptive tokenization~\cite{tome,evit,dynamicvit,tokenlearner}, while intra-LLM methods such as FastV~\cite{fastv}, DyToK~\cite{dytok}, PyramidDrop~\cite{pyramiddrop}, SparseVLM~\cite{sparsevlm}, and HiDrop~\cite{hidrop} prune tokens after partial multimodal reasoning. Recent adaptive methods further exploit visual complexity or input complexity~\cite{langdc,vico,tao2025dycoke,ffs,li2025divide}. These methods improve efficiency, but most of them allocate compression at the token or frame level without an explicit query-conditioned budget policy across the full video.

Query-aware frame and token selection methods condition visual processing on language input. Image-level saliency methods such as CROP~\cite{crop}, and video methods such as LGTTP~\cite{lgttp}, PruneVid~\cite{prunevid}, QTSplus~\cite{li2025seeing}, GenS~\cite{gens}, and VideoTree~\cite{videotree}, show that a subset of frames often dominates the answer. Router-based VLM designs such as long-tailed distribution-aware routing~\cite{ltdr} mainly focus on token-to-expert or mixture-of-experts behavior. In contrast, VideoRouter couples policy-level query routing with frame-level relevance estimation, enabling budgeted allocation between broad temporal coverage and high-resolution preservation.

%% file: sections/03_method.tex
\section{Method}
\label{sec:method}

Given a video $V=\{f_t\}_{t=1}^{T}$ and a question $q$, our goal is to minimize visual-token cost while preserving answer quality under a fixed context budget. Long videos contain substantial redundancy, but the evidence needed for a query is often distributed unevenly across time. VideoRouter addresses this by decomposing compression into two decisions: a policy-level decision that selects the dominant allocation policy, and a frame-level decision that estimates where high-fidelity visual evidence should be preserved. The final compressed visual sequence is inserted into the LLM prompt by replacing image placeholders, followed by standard autoregressive decoding.

\subsection{VideoRouter Architecture}
\label{sec:method:arch}

\begin{figure}[t]
\centering
\includegraphics[width=\textwidth]{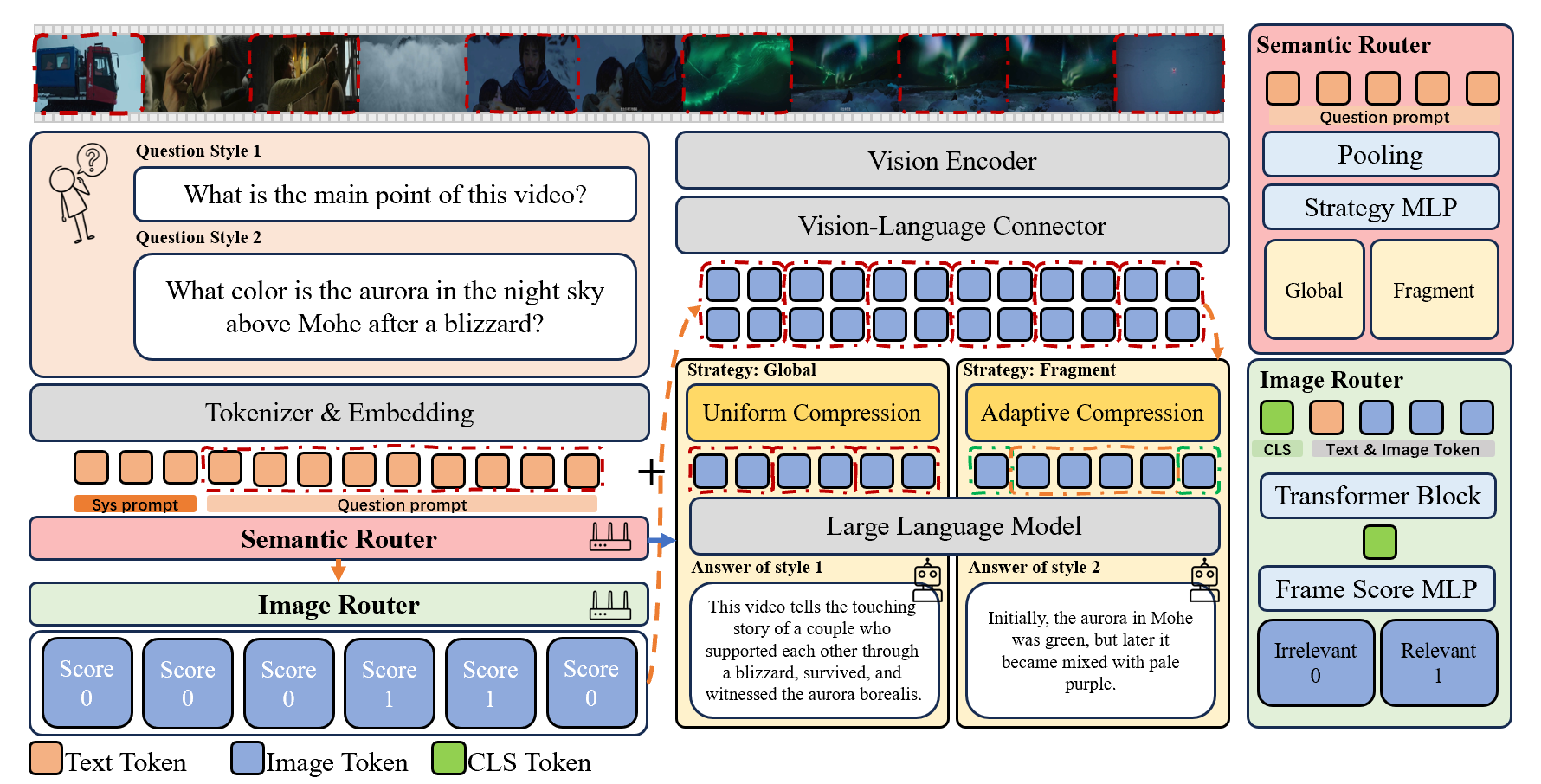}
\caption{Overview of the VideoRouter framework. The Semantic Router predicts the dominant allocation policy and selects either broad Uniform Compression or Adaptive Compression. Under Adaptive Compression, the Image Router's frame-level scores drive token allocation dynamically.}
\label{fig:method}
\end{figure}

\textbf{Backbone and Router Feature Extractor.} The model architecture consists of the open-source InternVL3 backbone and dual routers. For each frame, the vision encoder produces patch features and projects them into the LLM hidden space. To keep the routing mechanism lightweight and avoid introducing heavy additional parameters, VideoRouter reuses the early language layers as the router feature extractor. Concretely, the first four decoder layers are copied to a shadow stack $\mathcal{E}_{1:4}$, and the router features are computed as:
\begin{equation}
H_q = \mathcal{E}_{1:4}(\mathrm{Emb}(q)), \qquad
H_{v,q}^{(t)} = \mathcal{E}_{1:4}([Z_t;\mathrm{Emb}(q)]),
\end{equation}
where $Z_t$ is the frame-$t$ visual token sequence after vision projection, and $[\cdot;\cdot]$ denotes concatenation. This design leverages the multimodal fusion capabilities already present in the early layers of the LLM.

\textbf{Semantic Router.} The Semantic Router acts as a lightweight policy classifier over the question text to determine the dominant visual-token allocation policy. It first applies a Pooling operation to the token features $H_q$ with attention masking to aggregate the global query context. Then, it applies a Strategy MLP with LayerNorm and GELU activation, supervised by a cross-entropy loss:
\begin{equation}
p_s = g_s\!\left(\mathrm{Pooling}(H_q)\right), \qquad \hat{y}_s=\arg\max_c p_s(c), \qquad y_s\in\{0,1\},
\end{equation}
where $p_s(c)$ is the predicted probability of allocation policy $c$, $y_s=0$ denotes a \emph{Global} allocation policy, and $y_s=1$ denotes a \emph{Fragment} allocation policy. These labels specify how the token budget should be spent rather than imposing an exhaustive taxonomy of question semantics. If $\hat{y}_s=0$, VideoRouter applies Uniform Compression to preserve broad temporal coverage; if $\hat{y}_s=1$, it routes the visual sequence to Adaptive Compression, where the Image Router's per-frame scores dictate selective preservation of relevant frames.

\textbf{Image Router.} For adaptive allocation, the model needs to identify which frames contain evidence that should retain higher visual fidelity. Each frame is fused with the question tokens (Text \& Image Token) and encoded by the shadow stack $\mathcal{E}_{1:4}$. A lightweight Transformer Block with a learnable [CLS] token is then applied to capture the cross-modal interactions, followed by a Frame Score MLP to predict the per-frame relevance:
\begin{equation}
p_t = g_v\!\left(H_{v,q}^{(t)}\right), \qquad \hat{y}_t=\mathbb{I}[p_t>0.5], \qquad y_t\in\{0,1\}.
\end{equation}
where $p_t$ is the predicted relevance probability for frame $t$. In our implementation, $y_t=1$ indicates a Relevant frame that should retain high fidelity, while $y_t=0$ indicates an Irrelevant frame that can undergo aggressive compression.

\subsection{Inference}
\label{sec:inference}

The primary objective during inference is to execute the dual-router's decisions while strictly adhering to the LLM's context window. To achieve this, we formulate the post-routing compression as a budget-constrained optimization problem, ensuring zero out-of-memory errors and predictable latency.

\textbf{Dynamic Vision Budget Calculation.} 
Before processing the video, VideoRouter dynamically calculates the available visual token budget $B_{\mathrm{img}}$. Given the model's maximum context length $L_{\mathrm{max}}$, the text prompt length $L_{\mathrm{text}}$, and the reserved generation length $L_{\mathrm{gen}}$, the budget is defined as:
\[
B_{\mathrm{img}}=L_{\mathrm{max}}-L_{\mathrm{text}}-L_{\mathrm{gen}}-\epsilon
\]
where $\epsilon$ is a small safety margin. This dynamic budget strictly dictates the subsequent frame dropping and pooling strategies.

\textbf{Priority-Aware Frame Pooling.}
The Semantic Router's prediction $\hat{y}_s$ dictates the budget allocation policy. Let $N$ be the raw visual tokens per frame. Full pseudocode is provided in Appendix~\ref{app:algorithm}.

For the \emph{Global} allocation policy ($\hat{y}_s=0$), broad temporal coverage is prioritized. We apply a uniform spatial pooling scale $s_g=2$ to all frames. If the total token count still exceeds $B_{\mathrm{img}}$, VideoRouter safely falls back to uniform temporal subsampling.

For the \emph{Fragment} allocation policy ($\hat{y}_s=1$), token allocation is dynamic and guided by the Image Router's relevance scores $\hat{y}_t$. This policy can preserve multiple temporally distant evidence frames, so it is not limited to single-event questions. We adopt a two-tier resolution system: a high-fidelity scale $s_1=1$ for critical frames ($\hat{y}_t=1$) and an aggressive scale $s_0=4$ for irrelevant frames ($\hat{y}_t=0$). The framework allocates available token capacity following a strict priority hierarchy:

\begin{enumerate}[leftmargin=2em]
    \item \textit{Guarantee Critical Context:} Prioritize preserving all high-relevance frames at the high-fidelity scale $s_1$.
    \item \textit{Provide Background Context:} Allocate any remaining budget to low-relevance frames at scale $s_0$ to maintain a coarse temporal understanding.
    \item \textit{Strategic Discarding:} Under extreme budget constraints, discard low-priority frames first. If the budget cannot accommodate all critical frames, uniformly sample the high-priority set to ensure vital evidence still spans the timeline.
\end{enumerate}
The mathematical boundaries for these scenarios and the sampling execution are formalized in Algorithm~\ref{alg:budget_allocation} in the appendix.

\textbf{Sequence Reconstruction.} 
After spatial pooling and temporal dropping, the resulting sequence $Z_{\mathrm{final}}$ yields a highly condensed, variable-length token distribution per frame. We inject these refined tokens back into the multi-modal text prompt by replacing the designated image context placeholders. This dynamically compacted sequence is finally processed by the LLM backbone for standard autoregressive generation.

%% file: sections/04_supervision_data.tex
\section{Supervision and Data Construction}
\label{sec:supervision}

The routers in VideoRouter are trained to make allocation decisions rather than to answer benchmark questions directly. This section summarizes the supervision signals and training schedule; full construction details are provided in Appendix~\ref{app:data}.

\subsection{Data Construction}
\label{sec:supervision:data}

\begin{figure}[t]
\centering
\includegraphics[width=\textwidth]{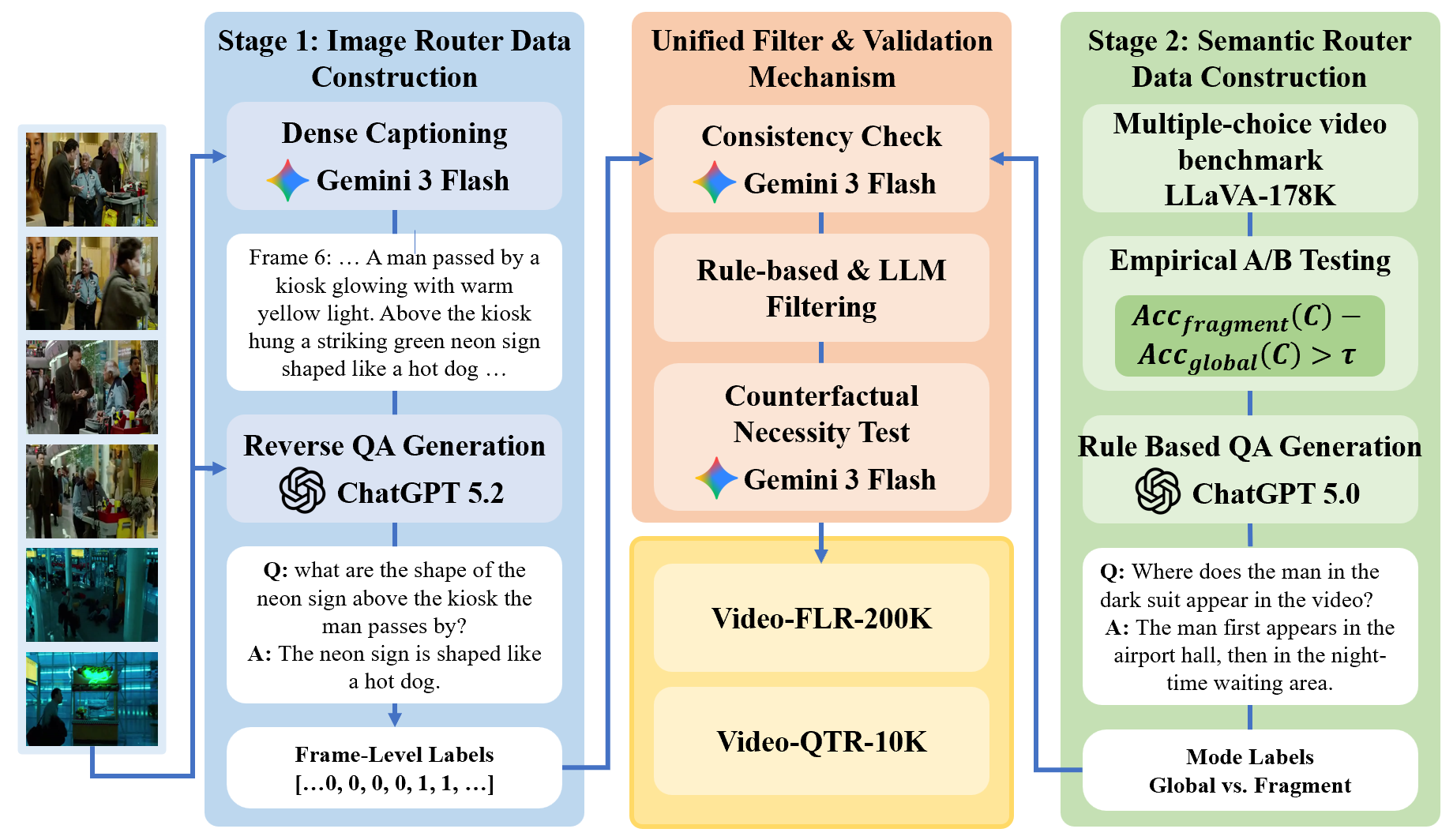}
\caption{Overview of the data construction pipeline. Stage 1 generates minimal sufficient frame-level labels for the Image Router, while Stage 2 derives Global vs. Fragment allocation-policy labels for the Semantic Router. All generated data undergoes a rigorous unified validation mechanism to ensure high fidelity.}
\label{fig:data_pipeline}
\end{figure}

Training VideoRouter requires two supervision signals: policy-level allocation labels for the Semantic Router and frame-level relevance labels for the Image Router. We construct Video-FLR-200K by identifying the minimal sufficient frames needed to answer generated reverse-QA pairs, and construct Video-QTR-10K by assigning each query category to the allocation policy that performs better in an empirical A/B probing benchmark. The raw visual materials are uniformly sampled from the open-source LLaVA-178K training dataset to maintain domain diversity. For a question category $C$, we assign the Fragment allocation label ($y_s=1$) if:
\begin{equation}
    Acc_{\text{fragment}}(C) - Acc_{\text{global}}(C) > \tau
\end{equation}
where $\tau$ is a predefined accuracy margin threshold; otherwise it receives the Global allocation label. All generated samples pass consistency, logical-coherence, and counterfactual-necessity filters before being included in the final datasets. A held-out label-quality audit is provided in Appendix~\ref{app:label_audit}.

\subsection{Multi-Stage Training}
\label{sec:supervision:training}

\textbf{Stage 1: Image Router Training.} We first freeze the vision encoder, projection layer, and LLM backbone, and train only the feature extractor and Image Router on Video-FLR-200K with binary frame-relevance supervision:
\begin{equation}
\mathcal{L}_{\mathrm{img}}=-\frac{1}{T}\sum_{t=1}^{T}\left[y_t\log(p_t)+(1-y_t)\log(1-p_t)\right]
\end{equation}
where $y_t$ is the frame-relevance label and $p_t$ is the predicted relevance probability.

\textbf{Stage 2: Semantic Router Training.} We then freeze the backbone, extractor, and Image Router, and train only the Semantic Router on Video-QTR-10K with cross-entropy loss:
\begin{equation}
\mathcal{L}_{\mathrm{sem}}=-\sum_{c\in\{0,1\}}y_{s,c}\log(p_s(c))
\end{equation}
where $c\in\{0,1\}$ denotes the allocation policy and $p_s(c)$ is the predicted probability of policy $c$.

\textbf{Stage 3: Joint Fine-Tuning.} Finally, we freeze the vision encoder, unfreeze the LLM backbone, and jointly align the compressed visual sequence with answer generation:
\begin{equation}
\mathcal{L}=\mathcal{L}_{\mathrm{LM}}+\lambda_s\mathcal{L}_{\mathrm{sem}}+\lambda_v\mathcal{L}_{\mathrm{img}}
\end{equation}
where $\mathcal{L}_{\mathrm{LM}}$ is the auto-regressive language modeling loss and $\lambda_s=\lambda_v=0.01$ in our implementation.

%% file: sections/05_experiments.tex
\section{Experiments}
\label{sec:experiments}

\subsection{Experimental Settings}
\label{sec:experiments:settings}

\textbf{Evaluations and Inference Setup.} 
We evaluate VideoRouter on several comprehensive video understanding benchmarks designed to assess long-form temporal reasoning capabilities. These include Video-MME~\cite{videomme} (spanning short, medium, and long videos up to an hour), LongVideoBench~\cite{longvideobench} (tailored for extended temporal reasoning), MLVU~\cite{mlvu}, LVBench, and EgoSchema (focusing on fine-grained spatiotemporal understanding). 

VideoRouter is built on the InternVL3-8B architecture with input frames resized to $448 \times 448$. We report two evaluation protocols. First, for controlled comparisons, all budgeted methods start from the same 64 sampled frames and share the same maximum visual-token budget $B_{\mathrm{img}}=12{,}288$; we report the actual post-compression tokens used by each method. This isolates the contribution of the allocation strategy from the number of observed frames. Second, for comparison with standard budgeted baselines, InternVL3* dynamically samples frames according to the same context budget and uses an average of 38 frames. VideoRouter samples 64 frames and fits them into the same maximum budget via dynamic dual-routing compression. The safety margin $\epsilon$ is set to 100 tokens in the dynamic budget calculation to reserve sufficient space for answer generation.

\textbf{Implementation Details.}
All controlled VideoRouter experiments use 4 NVIDIA H20 GPUs with DeepSpeed ZeRO-2. We follow the three-stage curriculum in Sec.~\ref{sec:supervision:training}: Image Router pre-training, Semantic Router training, and one epoch of joint fine-tuning. Full hyperparameters are in Appendix~\ref{app:implementation}.

\subsection{Main Results}
\label{sec:experiments:results}

\input{tables/main_results}

Table~\ref{tab:long_video_results} compares VideoRouter with proprietary MLLMs, open-source MLLMs, and token-compression methods. VideoRouter improves over the corresponding base backbones and remains competitive with specialized compression baselines; Qwen-specific efficiency diagnostics are in Appendix~\ref{app:qwen_adaptation}.

Table~\ref{tab:fair_comparison} isolates the allocation strategy under a controlled InternVL3-8B setting with the same 64 sampled frames and maximum token budget. VideoRouter outperforms uniform pooling, CLIP relevance pooling, and data-only fine-tuning, indicating that gains primarily come from query-conditioned budget allocation rather than data exposure alone; protocols are in Appendix~\ref{app:baseline_protocols}.

\input{tables/fair_comparison}

Despite aggressive token reduction, VideoRouter often outperforms dense or uniformly compressed baselines by preserving high-fidelity details on evidence frames while retaining coarse context elsewhere.

\subsection{Exploratory Experiment}
\label{sec:experiments:exploratory}

\textbf{Token Budget, Sampling Frames, and Compression Scales.}
VideoRouter's core advantage lies in dynamic token allocation (Algorithm~\ref{alg:budget_allocation}). To clarify this mechanism, we analyze the token budget $B_{\mathrm{img}}$, initial sampled frame count $T$, and compression scales $s_g, s_0, s_1$ on LongVideoBench. Fig.~\ref{fig:ablation_budget_depth}(a) summarizes the main trend, while the full budget table is provided in Appendix~\ref{app:budget}.

\begin{figure}[t!]
\centering
\includegraphics[width=\textwidth]{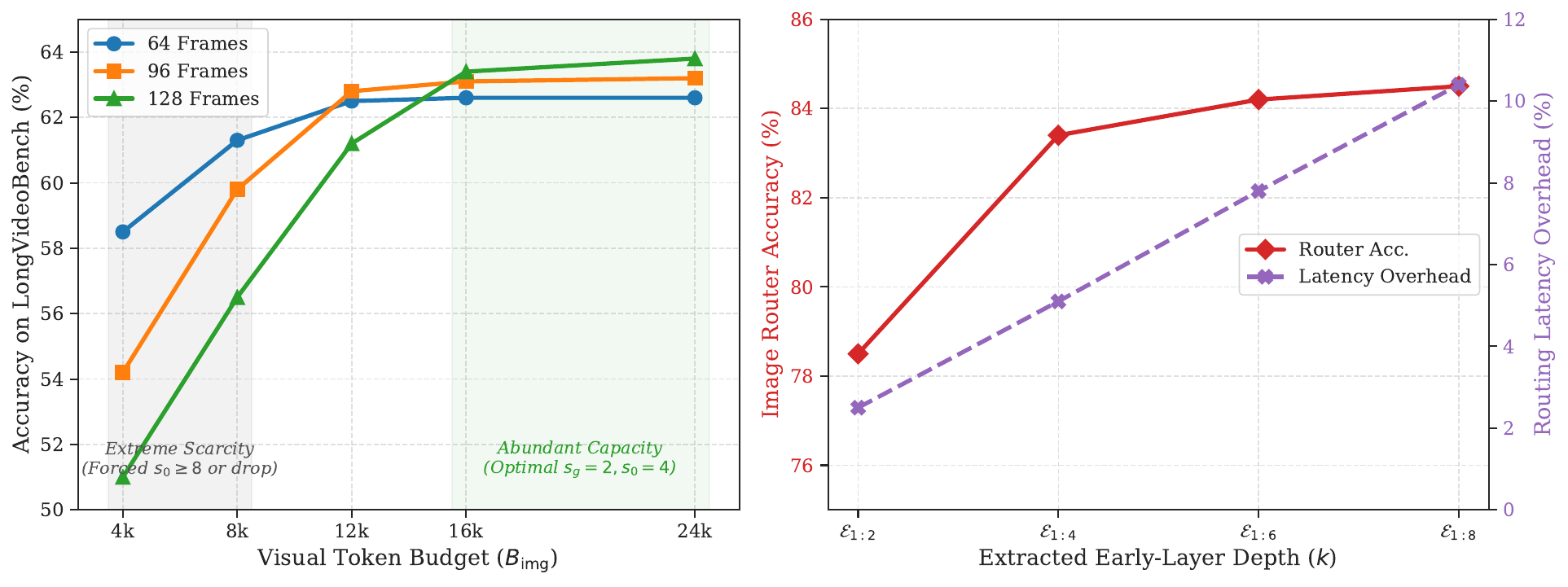}
\caption{(a) Unified analysis on LongVideoBench: The interplay between Token Budget, Sampled Frames, and dynamic compression scales. (b) The impact of the extracted early-layer depth ($k$) on Image Router accuracy and relative latency overhead.}
\label{fig:ablation_budget_depth}
\end{figure}

Under extreme token budget scarcity (4K--8K tokens), 64-frame sampling outperforms 128-frame dense sampling significantly. Forcing 128 frames into this budget triggers hyper-aggressive compression ($s_0 \geq 8$) on irrelevant frames and critical frame loss, damaging fine-grained spatial evidence. In contrast, 64-frame initialization maintains reasonable scales ($s_g=4, s_0=8$) without critical frame loss, preserving visual fidelity and achieving 61.3\% accuracy at 8K budget.

With abundant budget (16K--24K tokens), the trend reverses. The 64-frame configuration reaches a 62.6\% performance ceiling due to insufficient temporal coverage, while the 128-frame setup fully utilizes the budget to maintain optimal scales ($s_g=2, s_0=4$), retaining high spatial fidelity across broader temporal context and hitting a 63.8\% accuracy peak.

\textbf{Routing Effectiveness and Early-Layer Depth.} 
To evaluate dual-routing modules and justify default extraction depth $k=4$, we conduct quantitative assessments on dedicated test sets. The Semantic Router achieves 98.2\% accuracy on 2350 test samples, demonstrating strong capacity to infer a query's temporal scope from linguistic cues alone. The Image Router reuses early LLM layers $\mathcal{E}_{1:k}$ to score frame relevance, revealing a critical trade-off between extraction depth, accuracy on 2,500 samples, and latency overhead.

As illustrated in Fig.~\ref{fig:ablation_budget_depth}(b), two layers $\mathcal{E}_{1:2}$ yield sub-optimal 78.5\% accuracy, as superficial layers lack sufficient cross-modal semantic alignment for precise frame-to-text relevance. Increasing depth to $k=4$ boosts accuracy to 83.4\%, confirming these layers effectively capture multimodal cues for retaining critical frames. Further deepening to 6 or 8 layers brings marginal returns plateauing at 84.5\%, while routing latency overhead rises linearly from 5.1\% to over 10.4\%. Thus, $\mathcal{E}_{1:4}$ hits the optimal Pareto frontier, delivering robust high-precision routing with minimal computational footprint.

\subsection{Ablation Studies}
\label{sec:experiments:ablation}

\textbf{Effectiveness of Dual-Router Components.}
To evaluate the individual contributions of the Semantic and Image Routers, we conduct ablation studies on Video-MME. All variants maintain a unified token budget of $B_{\mathrm{img}}=12{,}288$, using fixed global scales of $s_g=2$ for Global queries and $s_0=4$ for irrelevant frames in Fragment queries.

\input{tables/ablation_studies}

The full model outperforms all variants in Table~\ref{tab:ablation_combined}(a). Removing the Image Router prevents fine-grained redundancy filtering, while removing the Semantic Router can over-apply adaptive allocation to holistic queries; thus policy prediction and frame-level relevance scoring are complementary.

\textbf{Necessity of Joint Fine-Tuning.}
Table~\ref{tab:ablation_combined}(b) validates the training curriculum. Skipping joint fine-tuning leaves the LLM poorly aligned with compressed visual tokens, while training without isolated router pre-training weakens routing supervision.

\textbf{Efficiency and Overhead Analysis.} VideoRouter adds a shadow-stack router pass, but Table~\ref{tab:latency_breakdown} shows that reduced LLM prefill more than offsets this overhead, lowering TTFT and peak memory.

\input{tables/latency_breakdown}

%% file: tables/main_results.tex
\begin{center}
\captionof{table}{Performance comparison on long-video benchmarks. We report accuracy (\%) on Video-MME, LongVideoBench, MLVU, LVBench, and EgoSchema. Results above the InternVL3 block are reported from prior work under their original settings, except Qwen2.5-VL + VideoRouter, which is our cross-backbone adaptation under the same budgeted allocation protocol. Bold method names mark our VideoRouter adaptations; the controlled InternVL3 comparison is provided in Table~\ref{tab:fair_comparison}.}
\label{tab:long_video_results}
\small
\resizebox{\textwidth}{!}{
\begin{tabular}{l cc ccccc}
\toprule
\multirow{2}{*}{\textbf{Model}} & \multicolumn{2}{c}{\textbf{Video-MME}} & \multirow{2}{*}{\textbf{LongVideoBench}} & \multirow{2}{*}{\textbf{MLVU}} & \multirow{2}{*}{\textbf{LVBench}} & \multirow{2}{*}{\textbf{EgoSchema}} \\
\cmidrule(lr){2-3}
& \textbf{Long} & \textbf{Overall} & & & & \\
Average Duration & 2386s & 1010s & 473s & 651s & 4101s & 180s \\
\midrule
\multicolumn{7}{c}{\textit{Proprietary MLLMs}} \\
\midrule
GPT-4V~\cite{gpt4v} & 53.5 & 59.9 & 59.1 & 49.2 & - & 55.6 \\
GPT-4o~\cite{gpt4o} & 65.3 & 71.9 & 66.7 & 64.6 & 30.8 & 72.2 \\
\midrule
\multicolumn{7}{c}{\textit{Open-source MLLMs}} \\
\midrule
Oryx-1.5~\cite{liu2024oryx} & - & 58.8 & 56.3 & 67.5 & - & - \\
MiniCPM-v2.6~\cite{yao2024minicpm} & 51.8 & 60.9 & 54.9 & 37.3 & - & - \\
mPLUG-Owl3~\cite{ye2024mplugowl3} & 50.1 & 59.3 & 52.1 & 63.7 & - & - \\
NVILA~\cite{lin2024nvila} & 54.8 & 64.2 & - & 70.1 & - & - \\
LLaVA-Video~\cite{llavavideo} & - & 63.3 & 58.2 & 70.8 & 41.5 & 57.3 \\
Video-XL~\cite{jiang2024videoxl} & 49.2 & 55.5 & 49.5 & 64.9 & - & - \\
VideoLLaMA2~\cite{cheng2024videollama2} & - & 47.9 & - & 48.5 & - & 51.7 \\
Video-CCAM~\cite{li2024videoccam} & 46.7 & 53.2 & - & 58.5 & - & - \\
Kangaroo~\cite{kangaroo} & 46.7 & 56.0 & 54.8 & 61.0 & 39.4 & 62.7 \\
LongVA~\cite{zhang2024longva} & 46.2 & 52.6 & - & 56.3 & - & - \\
LongVILA~\cite{longvila} & - & 60.1 & 57.1 & - & - & - \\
LongVU~\cite{shen2024longvu} & - & 60.6 & - & 65.4 & - & - \\
\midrule
\multicolumn{7}{c}{\textit{Compression Methods}} \\
\midrule
LLaVA-OneVision~\cite{li2024llavaonevision} & - & 58.2 & 56.5 & 64.7 & - & 60.1 \\
\quad + FastV~\cite{fastv} & - & 57.3 & - & - & - & - \\
\quad + PruMerge~\cite{prumerge} & - & 52.9 & - & - & - & - \\
\quad + DyCoke~\cite{tao2025dycoke} & - & 59.5 & - & - & - & - \\
\quad + DyToK~\cite{dytok} & 49.0 & 59.8 & 58.3 & 49.7 & - & - \\
\quad + QTSplus~\cite{li2025seeing} & - & 52.9 & - & 60.9 & 36.7 & - \\
\quad + DToMA~\cite{dtoma} & - & 65.0 & 59.6 & 71.7 & - & 59.3 \\
Qwen2.5-VL~\cite{wang2025qwen25vl} & - & 65.1 & 56.0 & 70.2 & 45.3 & 65.0 \\
\quad + DIG~\cite{li2025divide} & 55.3 & - & 61.4 & 70.7 & - & - \\
\quad + FastV~\cite{fastv} & - & 55.7 & 53.2 & 36.8 & - & - \\
\quad + DyToK~\cite{dytok} & - & 60.8 & 57.9 & 44.1 & - & - \\
\quad + \textbf{VideoRouter} & 55.8 & 66.0 & 60.4 & 71.0 & 46.0 & 65.7 \\
\midrule
InternVL3~\cite{internvl} & - & 66.3 & 58.8 & 71.4 & - & - \\
InternVL3* & 54.7 & 64.3 & 56.9 & 70.2 & 41.3 & 62.7 \\
\textbf{VideoRouter (Ours)} & 56.0 & 66.5 & 61.9 & 72.1 & 45.5 & 65.1 \\
\bottomrule
\end{tabular}
}
\end{center}

%% file: tables/fair_comparison.tex
\begin{table}[t]
\centering
\caption{Controlled comparison on InternVL3-8B. The dense row is an uncompressed reference, while all budgeted rows start from the same 64 sampled frames where applicable and share the same maximum visual-token budget $B_{\mathrm{img}}=12{,}288$; the table reports the actual post-compression visual tokens used by each method. ``Data FT'' fine-tunes the baseline on the same router-supervision QA data without using any routing module.}
\label{tab:fair_comparison}
\small
\resizebox{\textwidth}{!}{
\begin{tabular}{l c c c c c}
\toprule
\textbf{Method} & \textbf{Frames} & \textbf{Visual Tokens} & \textbf{Video-MME} & \textbf{LongVideoBench} & \textbf{MLVU} \\
\midrule
InternVL3 Dense & 64 & 16,384 & 66.3 & 58.8 & 71.4 \\
InternVL3 Budget & $\sim$38 & 9,528 & 64.3 & 56.9 & 70.2 \\
\midrule
Uniform Pooling & 64 & 8,192 & 65.4 & 58.9 & 70.6 \\
CLIP Relevance Pooling & 64 & 7,936 & 65.8 & 59.7 & 71.0 \\
Data FT + Uniform Pooling & 64 & 8,192 & 65.2 & 58.1 & 70.7 \\
\textbf{VideoRouter} & \textbf{64} & \textbf{7,748} & \textbf{66.5} & \textbf{61.9} & \textbf{72.1} \\
\bottomrule
\end{tabular}
}
\end{table}

%% file: tables/ablation_studies.tex
\begin{table}[t]
    \caption{Ablation studies across multiple video benchmarks. (a) Evaluates the individual contributions of the routing modules. (b) Validates the necessity of distinct router pre-training and joint fine-tuning stages. LVB denotes LongVideoBench.}
    \label{tab:ablation_combined}
    \centering
    \begin{subtable}[t]{0.48\textwidth}
        \centering
        \caption{Module Effectiveness}
        \resizebox{\linewidth}{!}{
        \begin{tabular}{l cccc}
        \toprule
        \multirow{2}{*}{\textbf{Model}} & \multicolumn{2}{c}{\textbf{MME}} & \multirow{2}{*}{\textbf{LVB}} & \multirow{2}{*}{\textbf{MLVU}} \\
        \cmidrule{2-3}
        & \textbf{Long} & \textbf{Overall} & & \\
        \midrule
        InternVL3* & 54.7 & 64.3 & 56.9 & 70.2 \\
        w/o Image Router & 55.1 & 65.4 & 58.9 & 70.6 \\
        w/o Semantic Router & 55.5 & 65.6 & 61.2 & 69.8 \\
        \textbf{VideoRouter} & \textbf{56.0} & \textbf{66.5} & \textbf{61.9} & \textbf{72.1} \\
        \bottomrule
        \end{tabular}
        }
    \end{subtable}
    \hfill
    \begin{subtable}[t]{0.48\textwidth}
        \centering
        \caption{Training Strategy}
        \resizebox{\linewidth}{!}{
        \begin{tabular}{l cccc}
        \toprule
        \multirow{2}{*}{\textbf{Training Strategy}} & \multicolumn{2}{c}{\textbf{MME}} & \multirow{2}{*}{\textbf{LVB}} & \multirow{2}{*}{\textbf{Ego}} \\
        \cmidrule{2-3}
        & \textbf{Long} & \textbf{Overall} & & \\
        \midrule
        InternVL3* & 54.7 & 64.3 & 56.9 & 62.7 \\
        w/o Joint FT (Stage 3) & 54.5 & 64.9 & 57.1 & 64.4 \\
        w/o Pre-train (Stage 1 \& 2) & 55.1 & 65.4 & 60.2 & 64.6 \\
        \textbf{VideoRouter (Full)} & \textbf{56.0} & \textbf{66.5} & \textbf{61.9} & \textbf{65.1} \\
        \bottomrule
        \end{tabular}
        }
    \end{subtable}
\end{table}

%% file: tables/latency_breakdown.tex
\begin{table}[t]
\centering
\caption{End-to-end latency breakdown on InternVL3-8B with 64 sampled frames. Router time includes the shadow-stack forward pass and routing heads. TTFT is measured after video loading and includes visual encoding, routing, compression, and LLM prefill.}
\label{tab:latency_breakdown}
\small
\resizebox{\textwidth}{!}{
\begin{tabular}{l c c c c c c}
\toprule
\textbf{Method} & \textbf{Visual Tokens} & \textbf{Router (s)} & \textbf{Vision+Compress (s)} & \textbf{LLM Prefill (s)} & \textbf{TTFT (s)} & \textbf{Peak Mem. (GB)} \\
\midrule
InternVL3 Dense & 16,384 & -- & 1.7 & 6.7 & 8.4 & 36.5 \\
InternVL3 Budget & 9,528 & -- & 1.1 & 3.9 & 5.0 & 25.8 \\
Uniform Pooling & 8,192 & -- & 1.2 & 3.6 & 4.8 & 24.6 \\
CLIP Relevance Pooling & 7,936 & 0.2 & 1.3 & 3.4 & 4.9 & 24.3 \\
\textbf{VideoRouter ($s=2,4$)} & \textbf{7,748} & \textbf{0.5} & \textbf{1.2} & \textbf{3.0} & \textbf{4.7} & \textbf{23.9} \\
\textbf{VideoRouter ($s=4,8$)} & \textbf{5,262} & \textbf{0.5} & \textbf{1.2} & \textbf{2.5} & \textbf{4.2} & \textbf{22.4} \\
\bottomrule
\end{tabular}
}
\end{table}

%% file: sections/06_conclusion.tex
\section{Conclusion}
\label{sec:conclusion}

We presented VideoRouter, a query-adaptive framework for budgeted long-video token compression. By combining policy-level routing with frame-level relevance scoring, VideoRouter achieves up to a 67.9\% token reduction while maintaining or improving accuracy across major benchmarks. Current limitations include primary evaluation on InternVL3-8B, offline access to sampled frames, and possible residual bias from VLM/LLM-assisted supervision, despite our Qwen2.5-VL adaptation and label-audit procedures. Since biases in the base VLM or automatically constructed supervision may affect which visual evidence the routers preserve or discard, deployment should audit performance across domains and demographic groups.

%% file: appendix.tex
\section{Additional Data Construction Details}
\label{app:data}

Training VideoRouter requires high-quality supervision for two routing decisions: frame-level relevance for the Image Router and allocation-policy selection for the Semantic Router. Raw visual materials are uniformly sampled from the open-source LLaVA-178K training dataset to avoid benchmark leakage and to cover diverse visual domains.

\textbf{Image Router data.} We seek the minimal sufficient set of frames required to answer a specific query. Instead of heuristic sampling, we employ an inversion strategy. First, a frontier VLM densely captions uniformly sampled video frames, extracting fine-grained spatial and temporal details. Second, an LLM examiner generates reverse-QA pairs that depend on specific visual cues from a subset of frames. Finally, we test whether answering each generated question requires each individual frame; frames deemed essential receive a binary relevance label of 1.

\textbf{Semantic Router data.} To derive allocation-policy labels, we construct a probing benchmark with 2,500 multiple-choice questions covering diverse task categories. We compare two forced policies using the same InternVL3-8B backbone: an adaptive Fragment policy equipped with the trained Image Router and a Global policy using uniform frame pooling. A category receives the Fragment label if the adaptive policy improves over the Global policy by more than the margin $\tau$; otherwise it receives the Global label.

\textbf{Unified validation.} All generated samples pass three filters. First, a consistency check verifies that the designated visual input is sufficient to recover the target answer. Second, an LLM-based logical-coherence filter removes contradictions between the generated QA, dense captions, and allocation label. Third, a counterfactual necessity test masks labeled positive frames or forces the alternative allocation policy; labels are corrected or removed if the answer remains recoverable without the claimed evidence.

\begin{figure}[h]
\centering
\includegraphics[width=\textwidth]{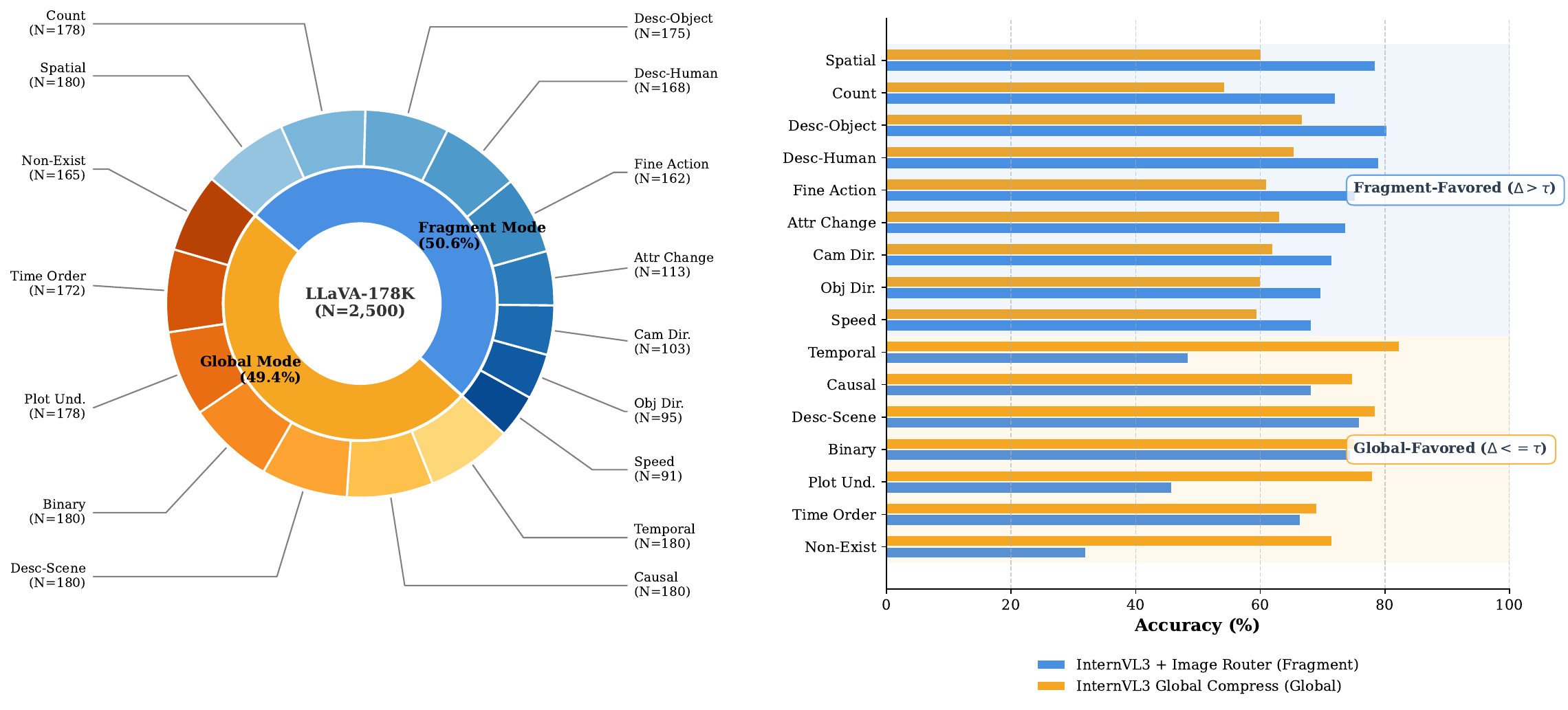}
\caption{Empirical A/B probing study for Semantic Router supervision. The left panel shows the distribution of sampled question categories, while the right panel compares forced Global and Fragment allocation policies. Categories with a sufficient Fragment-policy margin are assigned adaptive-allocation labels; the remaining categories are assigned Global allocation labels.}
\label{fig:app_ab_test}
\end{figure}

\section{Label Quality Audit}
\label{app:label_audit}

To estimate the quality of the generated supervision, we audit a held-out subset of 500 samples, including 250 policy-label examples from Video-QTR-10K and 250 frame-relevance examples from Video-FLR-200K. Each sample is independently checked by two annotators using the source video frames, generated question, answer, and assigned labels. For allocation-policy labels, annotators decide whether broad coverage or adaptive high-resolution preservation is more appropriate under a fixed budget. For frame-relevance labels, annotators mark whether each positive frame contains evidence necessary to answer the query.

The audit yields 94.0\% agreement with the generated allocation-policy labels and 88.4\% agreement with frame-relevance labels. Disagreements are concentrated in mixed queries requiring both broad temporal context and localized evidence, as well as visually ambiguous frames where multiple adjacent frames contain near-duplicate evidence. These results suggest that the validation pipeline substantially reduces teacher noise, while also motivating the use of a policy-level router rather than treating Global/Fragment as an exhaustive semantic taxonomy.

\section{Data Statistics and Split Integrity}
\label{app:data_stats}

Video-FLR-200K contains 200,664 QA instances with frame-level relevance annotations, and Video-QTR-10K contains 10,000 video-query pairs with allocation-policy labels. We reserve held-out subsets for router validation and label-quality auditing, and no benchmark validation or test videos are used during data construction. The raw visual materials are sampled from the training split of LLaVA-178K to provide broad visual diversity while avoiding contamination from Video-MME, LongVideoBench, MLVU, LVBench, and EgoSchema evaluation examples.

The generated labels are used only to supervise routing and allocation decisions. They do not provide direct benchmark answers at inference time. This separation is important because VideoRouter changes how visual evidence is represented under a budget, rather than adding task-specific answer memorization.

\section{Additional Baseline Protocols}
\label{app:baseline_protocols}

We describe the controlled baselines used in Table~\ref{tab:fair_comparison} and Table~\ref{tab:latency_breakdown}. Except for the dense uncompressed reference row, all controlled methods use InternVL3-8B, the same 64 sampled input frames where applicable, and the same maximum visual-token budget $B_{\mathrm{img}}=12{,}288$. The table reports actual post-compression visual tokens because some fixed pooling strategies do not consume the full maximum budget exactly.

\textbf{Uniform Pooling.} Uniform Pooling applies the same spatial pooling scale to every sampled frame, independent of the query. We use the strongest fixed scale that fits comfortably within the budget while preserving all 64 frames, yielding 8,192 visual tokens. We also tried using a weaker pooling scale to consume more tokens, but it exceeded the effective budget once text and generation reservations were included.

\textbf{CLIP Relevance Pooling.} CLIP Relevance Pooling computes frame-query similarity using a frozen CLIP-style image-text encoder. Frames are ranked by similarity; the highest-scoring frames retain higher spatial fidelity, while lower-scoring frames are aggressively pooled until the resulting sequence fits the maximum budget. This baseline uses no VideoRouter training data or early LLM features, and therefore tests whether a training-free relevance score is sufficient for budget allocation.

\textbf{Data FT + Uniform Pooling.} This baseline fine-tunes InternVL3-8B on the same router-supervision QA data used by VideoRouter but removes both routing modules at inference time. It then applies uniform pooling to all frames. This isolates the effect of data exposure from the effect of dynamic allocation, since it sees the same training examples but cannot adapt token allocation per query or frame.

\textbf{Latency measurement.} TTFT is averaged over the same evaluation subset after video loading. It includes visual encoding, optional routing, compression, and LLM prefill. Router time includes the shadow-stack forward pass and routing heads. The latency table is intended to account for the router overhead rather than reporting token reduction alone. All latency numbers are reported as single-sample inference averages on the same hardware configuration used in the main experiments.

\section{Cross-Backbone Adaptation on Qwen2.5-VL}
\label{app:qwen_adaptation}

To verify that VideoRouter is not tied to InternVL3-specific implementation details, we instantiate the same routing framework on Qwen2.5-VL. The adaptation keeps the allocation algorithm unchanged: the Semantic Router predicts the dominant allocation policy, the Image Router scores frame relevance, and the resulting frame scores drive the same priority-aware pooling rule. The only backbone-specific change is the router feature extractor, where we reuse the first four language layers of Qwen2.5-VL as the shadow stack and attach the same lightweight router heads.

Qwen2.5-VL uses dynamic-resolution visual tokenization~\cite{wang2025qwen25vl}. As an approximate image-frame intuition, its 14-pixel patching followed by a $2\times2$ spatial merge makes one final visual token correspond roughly to a $28\times28$ image region; for a $448\times448$ image, this would be about $16\times16=256$ spatial visual tokens before video-specific temporal packing and downstream budgeted pooling. The exact video token count is not fixed per frame and depends on dynamic resizing, temporal packing, and text/generation reservations, so we report actual post-compression visual tokens in Table~\ref{tab:app_qwen_adaptation}.

The cross-backbone experiment uses the same high-level evaluation protocol as the main budgeted setting: 64 sampled frames, a maximum visual-token budget of 12,288, and the same safety margin for text and generation tokens. We compare against Qwen2.5-VL baselines under the same input-frame setting and report actual post-compression visual tokens to make the efficiency comparison explicit. Table~\ref{tab:app_qwen_fair} mirrors the controlled InternVL3 comparison in the main paper, while Table~\ref{tab:app_qwen_adaptation} reports Qwen-specific efficiency and latency diagnostics.

\begin{center}
\captionof{table}{Controlled fair comparison using Qwen2.5-VL as the base model. The dense row is an uncompressed 64-frame reference, while all budgeted rows use the same 64 sampled input frames and the same maximum visual-token budget. The table reports actual post-compression visual tokens.}
\label{tab:app_qwen_fair}
\small
\resizebox{\textwidth}{!}{
\begin{tabular}{l c c c c c}
\toprule
\textbf{Method} & \textbf{Frames} & \textbf{Visual Tokens} & \textbf{Video-MME} & \textbf{LongVideoBench} & \textbf{MLVU} \\
\midrule
Qwen2.5-VL Dense & 64 & 16,384 & 65.5 & 56.8 & 70.5 \\
Qwen2.5-VL Budget & 64 & 9,680 & 65.1 & 56.0 & 70.2 \\
\midrule
Uniform Pooling & 64 & 8,192 & 65.3 & 56.9 & 70.3 \\
CLIP Relevance Pooling & 64 & 7,936 & 65.6 & 57.6 & 70.7 \\
Data FT + Uniform Pooling & 64 & 8,192 & 65.4 & 56.6 & 70.6 \\
\textbf{Qwen2.5-VL + VideoRouter} & \textbf{64} & \textbf{7,910} & \textbf{66.0} & \textbf{60.4} & \textbf{71.0} \\
\bottomrule
\end{tabular}
}
\end{center}

The Qwen controlled comparison follows the same pattern as the InternVL3 study: using the same frames and budget, training-free CLIP relevance improves over uniform pooling but remains below VideoRouter, while data-only fine-tuning does not recover the gains of query-conditioned allocation. The largest margin appears on LongVideoBench, consistent with the paper's central claim that dynamic evidence allocation is most useful when sparse key events must be preserved under a long-video budget.

\begin{center}
\captionof{table}{Controlled cross-backbone comparison using Qwen2.5-VL as the base model. Both rows use 64 sampled input frames and the same maximum visual-token budget; the table reports actual post-compression visual tokens, TTFT, peak memory, and representative benchmark accuracy.}
\label{tab:app_qwen_adaptation}
\small
\resizebox{\textwidth}{!}{
\begin{tabular}{l c c c c c c}
\toprule
\textbf{Method} & \textbf{Frames} & \textbf{Visual Tokens} & \textbf{TTFT (s)} & \textbf{Peak Mem. (GB)} & \textbf{Video-MME Overall} & \textbf{LongVideoBench} \\
\midrule
Qwen2.5-VL Budget Baseline & 64 & 9,680 & 5.3 & 26.1 & 65.1 & 56.0 \\
\quad + VideoRouter & 64 & 7,910 & 4.9 & 24.7 & 66.0 & 60.4 \\
\bottomrule
\end{tabular}
}
\end{center}

The comparison shows that VideoRouter reduces the actual visual-token count by 18.3\% relative to the Qwen2.5-VL budget baseline, lowers TTFT by 0.4s, and improves LongVideoBench by 4.4 points. This trend is consistent with the InternVL3 experiments: dynamic allocation is most beneficial on long-context benchmarks where preserving sparse high-value evidence matters more than uniformly retaining all visual tokens.

\section{Dynamic Allocation Algorithm}
\label{app:algorithm}

\begin{algorithm}[h]
\caption{Dynamic Token Allocation under Budget Constraint}
\label{alg:budget_allocation}
\textbf{Input:} Token budget $B_{\mathrm{img}}$, frames $V=\{f_t\}_{t=1}^T$, relevance scores $\{\hat{y}_t\}_{t=1}^T$, tokens per frame $N$, pooling scales $s_1, s_0$ \\
\textbf{Output:} Pooled visual token sequence $Z_{\mathrm{final}}$
\begin{algorithmic}[1]
\STATE Initialize $\mathcal{F}_1 \leftarrow \{f_t \mid \hat{y}_t = 1\}$, $\mathcal{F}_0 \leftarrow \{f_t \mid \hat{y}_t = 0\}$
\STATE Calculate $\mathcal{C}_1 \leftarrow |\mathcal{F}_1| \cdot (N/s_1^2)$, $\mathcal{C}_0 \leftarrow |\mathcal{F}_0| \cdot (N/s_0^2)$
\IF{$\mathcal{C}_1 + \mathcal{C}_0 \le B_{\mathrm{img}}$}
    \STATE $\mathcal{F}_{\mathrm{keep}} \leftarrow \mathcal{F}_1 \cup \mathcal{F}_0$ \hfill $\triangleright$ \textit{Sufficient budget: keep all}
\ELSIF{$\mathcal{C}_1 \le B_{\mathrm{img}}$}
    \STATE $k \leftarrow \lfloor (B_{\mathrm{img}} - \mathcal{C}_1) / (N/s_0^2) \rfloor$ \hfill $\triangleright$ \textit{Partial budget: sample redundant}
    \STATE $\mathcal{F}_0' \leftarrow \text{UniformSample}(\mathcal{F}_0, k)$
    \STATE $\mathcal{F}_{\mathrm{keep}} \leftarrow \mathcal{F}_1 \cup \mathcal{F}_0'$
\ELSE
    \STATE $k \leftarrow \lfloor B_{\mathrm{img}} / (N/s_1^2) \rfloor$ \hfill $\triangleright$ \textit{Extreme budget: sample critical}
    \STATE $\mathcal{F}_{\mathrm{keep}} \leftarrow \text{UniformSample}(\mathcal{F}_1, k)$
\ENDIF
\STATE $Z_{\mathrm{final}} \leftarrow [\text{MeanPool}(f_t,\, s_1 \text{ if } \hat{y}_t = 1 \text{ else } s_0) \text{ for } f_t \text{ in } V \text{ if } f_t \in \mathcal{F}_{\mathrm{keep}}]$
\STATE \textbf{return} $Z_{\mathrm{final}}$
\end{algorithmic}
\end{algorithm}

\section{Budget Sensitivity}
\label{app:budget}

Table~\ref{tab:budget_sensitivity} reports the complete LongVideoBench budget sweep. We render it here as a non-floating table to keep the result adjacent to the discussion in the appendix.

\begin{center}
\captionof{table}{Budget sensitivity on LongVideoBench accuracy (\%). Each column denotes a rounded available visual-token budget $B_{\mathrm{img}}$ after reserving text and generation tokens: 4K, 8K, 12K, 16K, and 24K correspond approximately to 4,096, 8,192, 12,288, 16,384, and 24,576 visual tokens. All methods use InternVL3-8B.}
\label{tab:budget_sensitivity}
\small
\resizebox{\textwidth}{!}{
\begin{tabular}{l c c c c c}
\toprule
\multirow{2}{*}{\textbf{Method}} & \multicolumn{5}{c}{\textbf{Visual-token budget $B_{\mathrm{img}}$}} \\
\cmidrule(lr){2-6}
& \textbf{4K} & \textbf{8K} & \textbf{12K} & \textbf{16K} & \textbf{24K} \\
\midrule
Uniform Pooling (64f) & 55.8 & 58.0 & 58.9 & 60.0 & 60.7 \\
CLIP Relevance Pooling (64f) & 56.9 & 59.2 & 60.3 & 61.1 & 61.5 \\
\textbf{VideoRouter (64f)} & \textbf{58.4} & \textbf{61.3} & 61.9 & 62.6 & 62.6 \\
\textbf{VideoRouter (128f)} & 54.9 & 59.8 & \textbf{62.4} & \textbf{63.4} & \textbf{63.8} \\
\bottomrule
\end{tabular}
}
\end{center}

\section{Additional Implementation Details}
\label{app:implementation}

All controlled InternVL3 VideoRouter experiments use InternVL3-8B with input frames resized to $448\times448$. Stage 1 trains the Image Router on Video-FLR-200K for 5 epochs with learning rate $5\times10^{-5}$, per-device batch size 16, and 4 gradient accumulation steps. Stage 2 trains the Semantic Router on Video-QTR-10K for 2 epochs with learning rate $1\times10^{-5}$, per-device batch size 4, and 4 gradient accumulation steps. Stage 3 jointly aligns routing and generation for 1 epoch with learning rate $5\times10^{-6}$, per-device batch size 4, and 4 gradient accumulation steps. Unless otherwise specified, we set the safety margin in the dynamic budget calculation to $\epsilon=100$ tokens.

\begin{table}[h]
\centering
\caption{Detailed hyperparameters for the three-stage training curriculum. The Global and Fragment scales denote the spatial pooling scales used during joint fine-tuning and inference to fit routed visual tokens into the predefined budget.}
\label{tab:app_hyperparameters}
\resizebox{0.9\textwidth}{!}{
\begin{tabular}{l ccc}
\toprule
\textbf{Hyperparameter} & \textbf{Stage 1} & \textbf{Stage 2} & \textbf{Stage 3} \\
& \textbf{Image Router} & \textbf{Semantic Router} & \textbf{Joint FT} \\
\midrule
\texttt{model\_max\_length} & 12288 & 12288 & 12288 \\
\texttt{num\_train\_epochs} & 5 & 2 & 1 \\
\texttt{batch\_size} (per device) & 16 & 4 & 4 \\
\texttt{gradient\_accumulation} & 4 & 4 & 4 \\
\texttt{learning\_rate} & \num{5e-5} & \num{1e-5} & \num{5e-6} \\
\texttt{weight\_decay} & 0.0 & 0.0 & 0.05 \\
\texttt{warm\_up\_ratio} & 0.03 & 0.03 & 0.03 \\
\texttt{lr\_scheduler\_type} & Cosine & Cosine & Cosine \\
\midrule
\texttt{freeze\_vision\_model} & True & True & True \\
\texttt{freeze\_language\_model} & True & True & False \\
\texttt{freeze\_Semantic\_Router} & True & False & False \\
\texttt{freeze\_Image\_Router} & False & True & False \\
\midrule
\texttt{dropout} & 0.0 & 0.0 & 0.0 \\
\texttt{global\_scale} ($s_g$) & -- & -- & 2 \\
\texttt{fragment\_scale} ($s_1, s_0$) & -- & -- & 1, 4 \\
\bottomrule
\end{tabular}
}
\end{table}

\section{Router Training Dynamics}
\label{app:router_training}

\begin{figure}[h]
\centering
\includegraphics[width=0.48\textwidth]{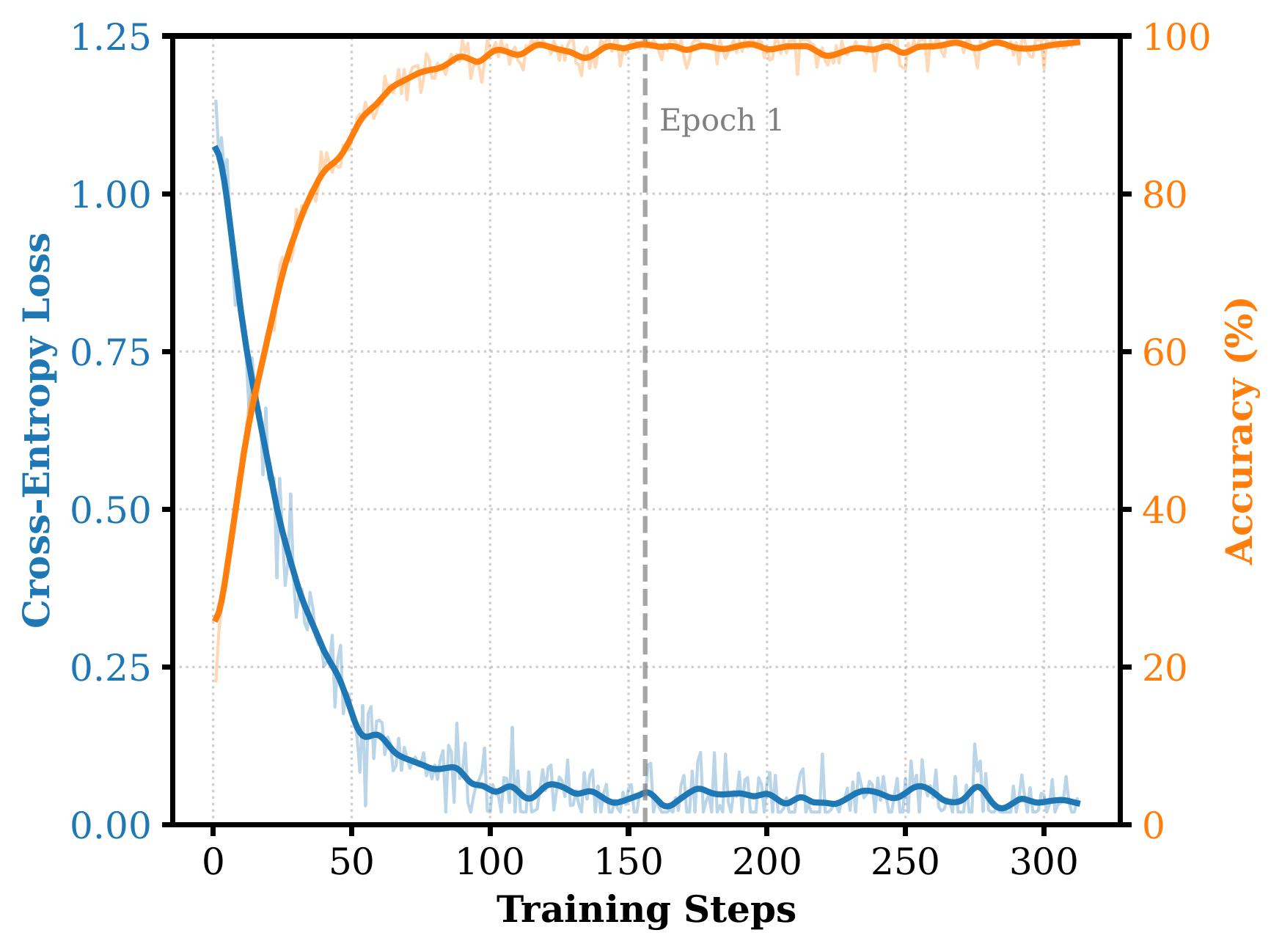}
\includegraphics[width=0.48\textwidth]{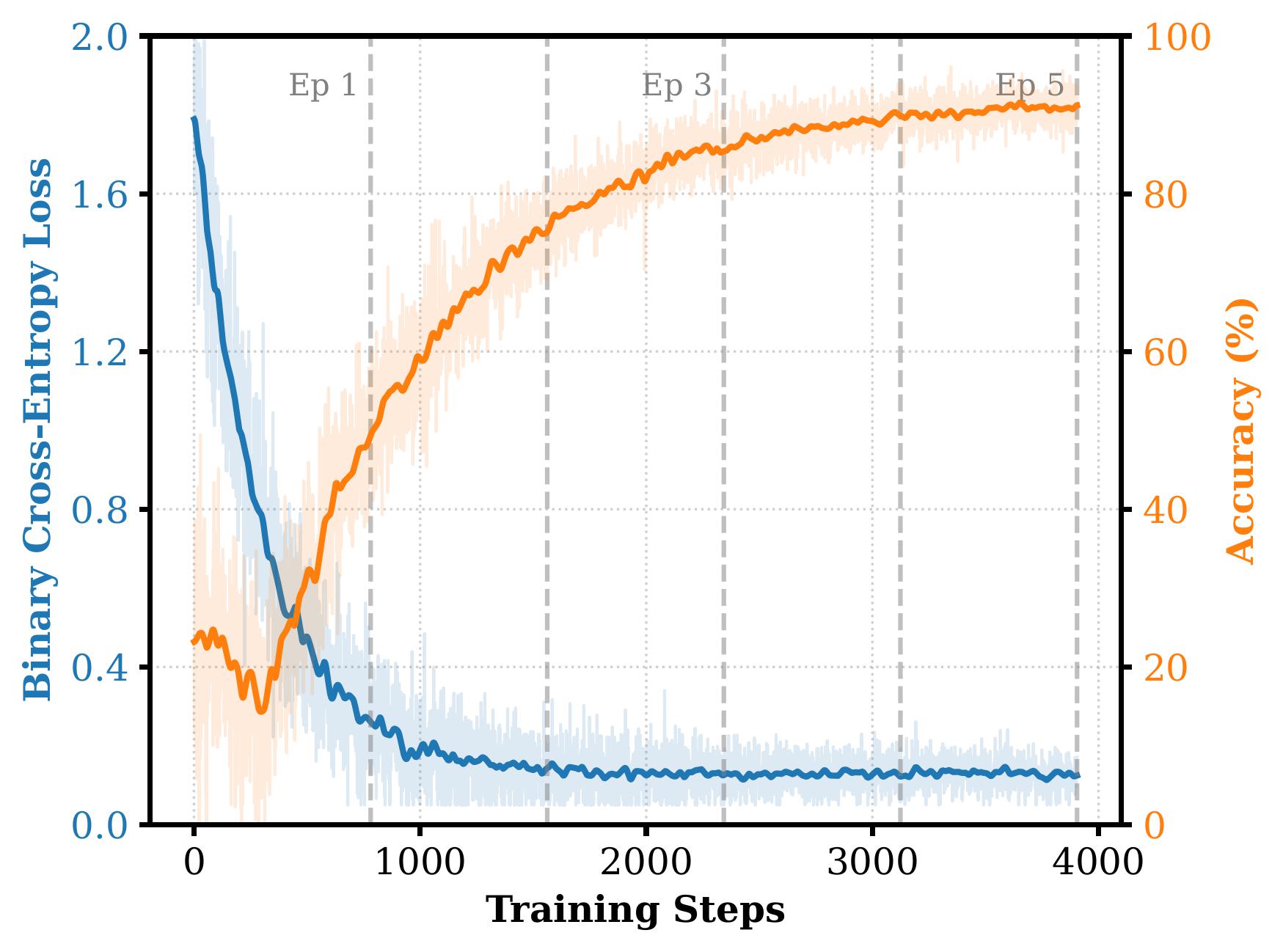}
\caption{Training curves for the Semantic Router and Image Router.}
\label{fig:app_router_training}
\end{figure}

Fig.~\ref{fig:app_router_training} shows that both routing modules converge stably. The Semantic Router converges rapidly, suggesting that early language features encode useful policy-level priors. The Image Router improves more gradually because frame-level relevance requires cross-modal alignment between visual evidence and the query.

\section{Prompt Templates}
\label{app:prompts}

\begin{figure}[h]
\centering
\includegraphics[width=\textwidth]{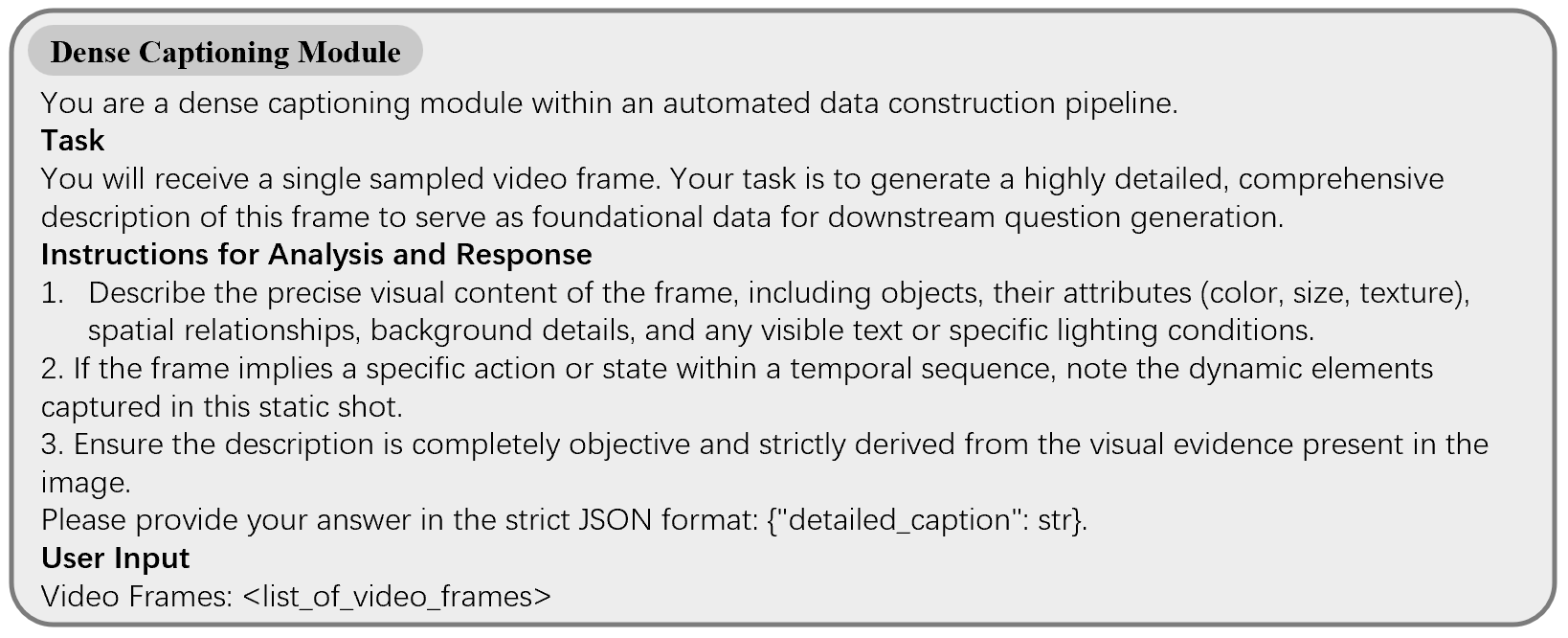}
\caption{Prompt template for dense captioning, designed to extract objective visual details from uniformly sampled frames.}
\label{fig:prompt_dense_caption}
\end{figure}

\begin{figure}[h]
\centering
\includegraphics[width=\textwidth]{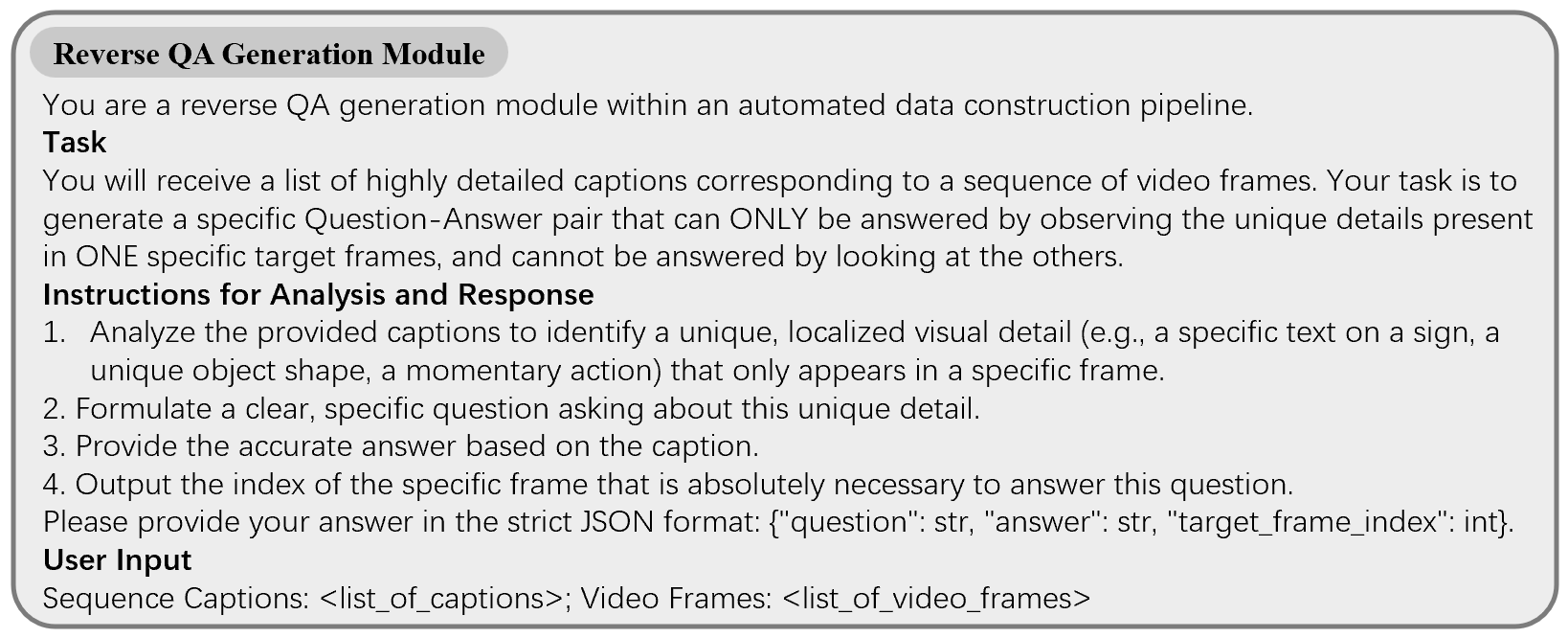}
\caption{Prompt template for reverse-QA generation, encouraging questions that rely on localized visual cues.}
\label{fig:prompt_reverse_qa}
\end{figure}

\begin{figure}[h]
\centering
\includegraphics[width=\textwidth]{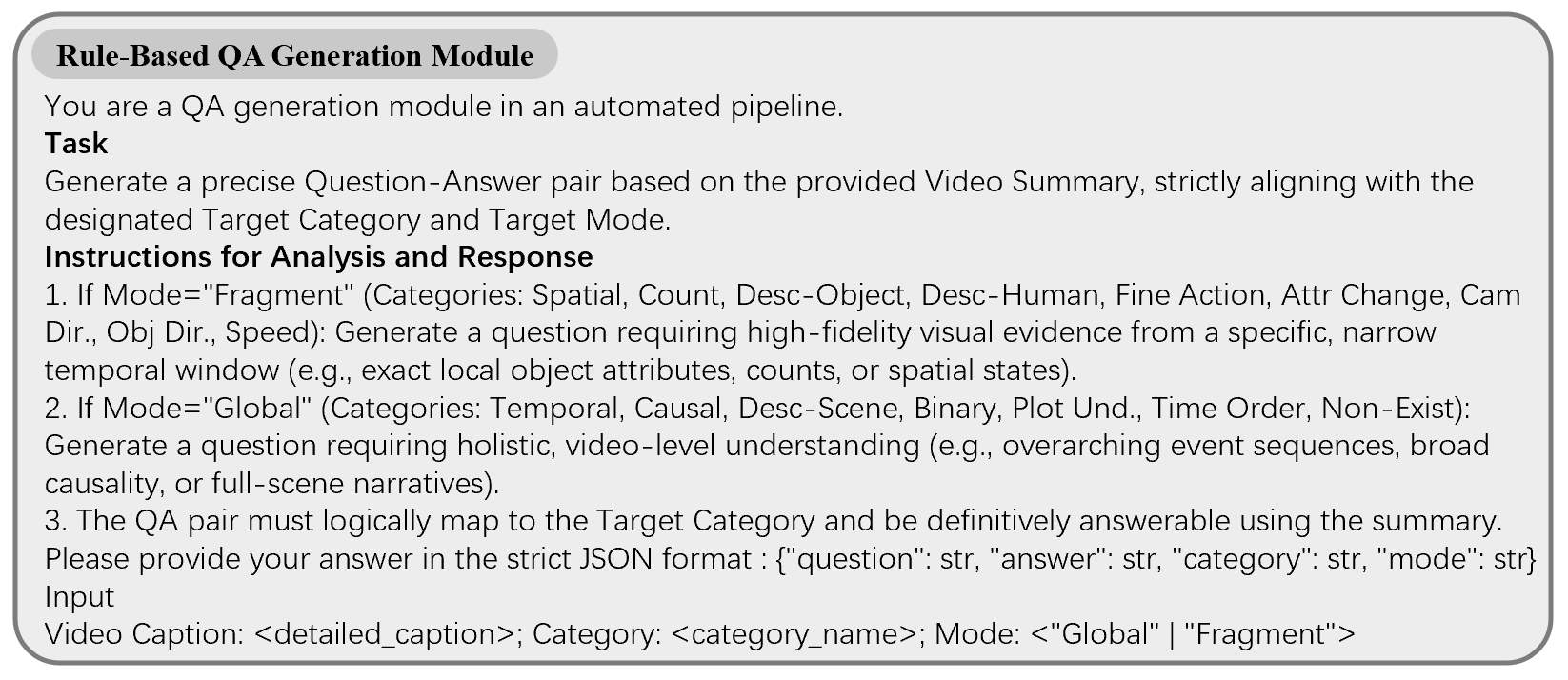}
\caption{Prompt template for rule-guided QA generation based on statistically selected allocation-policy categories.}
\label{fig:prompt_rule_qa}
\end{figure}

\begin{figure}[h]
\centering
\includegraphics[width=\textwidth]{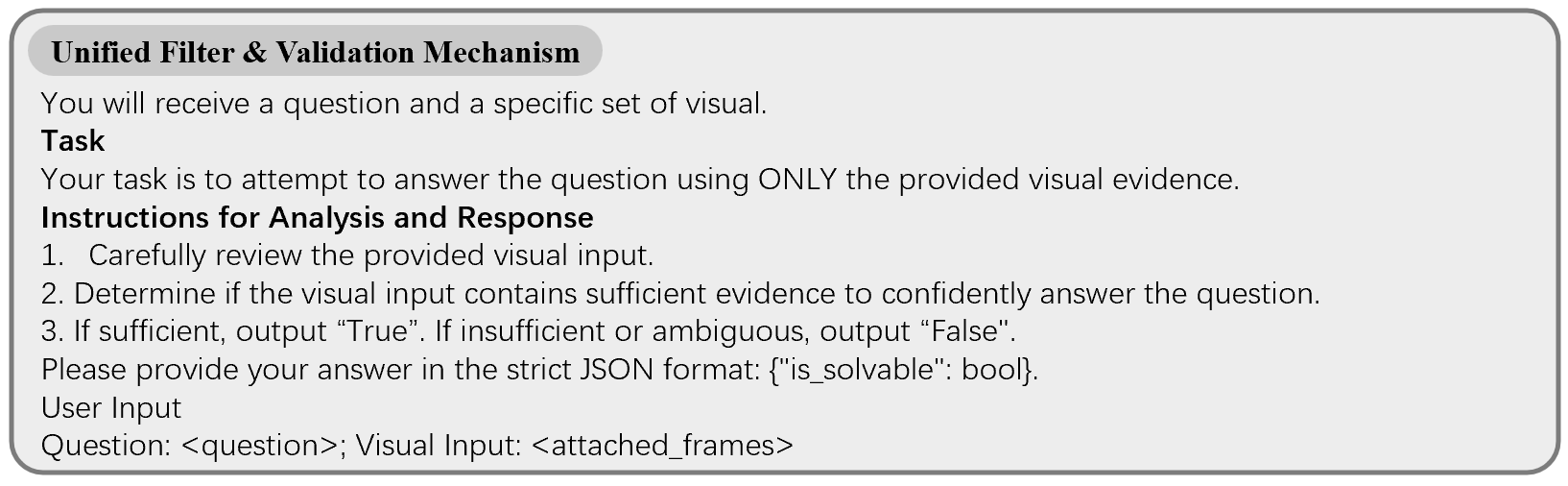}
\caption{Prompt template for consistency and counterfactual validation, ensuring visual sufficiency and necessity.}
\label{fig:prompt_consistency}
\end{figure}

\begin{figure}[h]
\centering
\includegraphics[width=\textwidth]{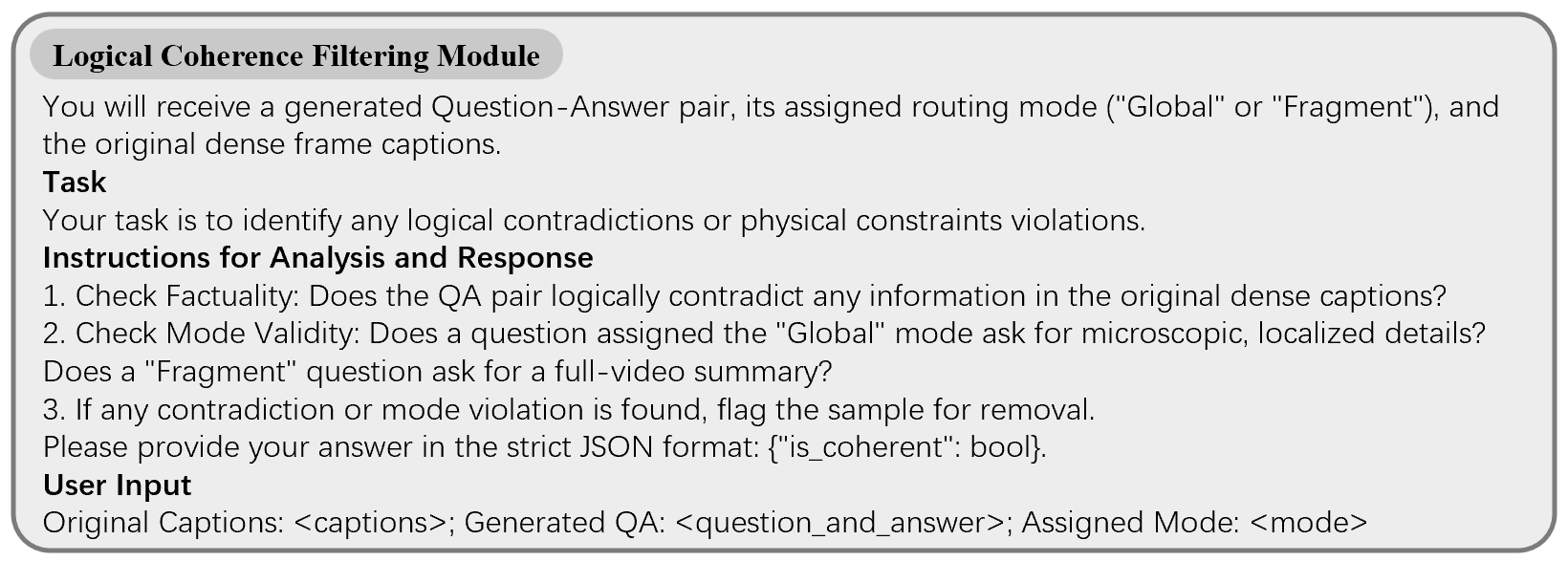}
\caption{Prompt template for logical-coherence filtering, identifying physical contradictions and allocation-policy violations.}
\label{fig:prompt_logical}
\end{figure}

Fig.~\ref{fig:prompt_dense_caption}--Fig.~\ref{fig:prompt_logical} show the prompt templates used by the automated data construction and validation pipeline.

